\documentclass[english]{revtex4-2}
\usepackage[T1]{fontenc}
\usepackage[latin9]{inputenc}
\usepackage{textcomp}
\usepackage{graphicx}
\usepackage{babel}
\begin{document}
\title{The artificial synesthete: Image-melody translations with variational
autoencoders}
\author{Karl Wienand}
\affiliation{Technische Universit\"at M\"unchen, Munich, Germany}
\affiliation{Deutsches Museum, Munich, Germany}
\author{Wolfgang M. Heckl}
\affiliation{Technische Universit\"at M\"unchen, Munich, Germany}
\affiliation{Deutsches Museum, Munich, Germany}

\begin{abstract}
This project presents a system of neural networks to translate between
images and melodies. Autoencoders compress the information in samples
to abstract representation. A translation network learns a set of
correspondences between musical and visual concepts from repeated
joint exposure. The resulting \textquotedblleft artificial synesthete\textquotedblright{}
generates simple melodies inspired by images, and images from music.
These are novel interpretation (not transposed data), expressing the
machine\textquoteright s perception and understanding. Observing the
work, one explores the machine\textquoteright s perception and thus,
by contrast, one\textquoteright s own.
\end{abstract}
\maketitle
Translating between visual arts and music using computers has always
woven together science, technology, and art. Technologists started
performing the first pioneering experiments with automated generation
of image and music in the 1950s and 60s \cite{noll_patterns_1962,hiller_musical_1958},
which evolved into an assistance to human artists \citet{xenakis_formalized_1992,taylor_when_2014,daudrich_algorithmic_2016,berg_composing_2009}.
Recent advances in hardware and algorithms made neural-networks-based
generation widely accessible for both research and art \citet{alvarez-melis_emotional_2017,briot_deep_2019,roche_autoencoders_2018,broad_autoencoding_2017,roberts_hierarchical_2018,larsen_autoencoding_2016,diaz-jerez_composing_2011,carnovalini_computational_2020,fernandez_ai_2013,goodfellow_generative_2014}.
Scientific and artistic interests also meet in bridging between the
expressions: music visualization---an idea rooted in early 20th century
art \citet{corra_abstract_1912,kandinsky_uber_1912,moritz_optical_2004}---is
mostly the purview of artists \citet{ox_curating_2006}, while much
of the image-to-music transformation is based on sonification, often
with a scientific-technological focus \citet{barrass_using_1999,dubus_systematic_2013,kramer_sonification_2010,hermann_theory_2011}.

One-to-one mappings between elements (for example pixels and notes)
are a simple, immediate, and common way to translate between image
and music. Since the transformation is bijective, simple algorithms
can use it to visualize music as well as play images. Pablo Samuel
Castro\textquoteright s jidiji experiment \citet{castro_jidiji:_nodate}
and Wolfgang M. Heckl\textquoteright s installation \emph{Atomare
Klangwelte} \citet{heckl_atomare_2006} are but two examples. However,
the musical results often sound random, because these algorithms blindly
transpose data. Neural networks, thanks to their considerable autonomy,
offer more generative power and potential for artificial creativity
\citet{saunders_towards_2012,jordanous_standardised_2012,colton_creativity_2008}.
A recent work \citet{muller-eberstein_translating_2019}, for example,
used variational autoencoder networks to generate novel (i.e. not
transposed) melodies from artworks, based on idiosyncratic audiovisual
correspondences. We present a system of neural networks, sketched
in Figure 1 constituting an \textquotedblleft artificial synesthete\textquotedblright{}
able to translate between simple melodies and pictures. The translation
uses a core of cross-modal correspondences to link musical and visual
concepts. This allow us to better comprehend how the machine understands
information and how it applies, reinterprets, or breaks the rules.

\begin{figure}[th]
\includegraphics{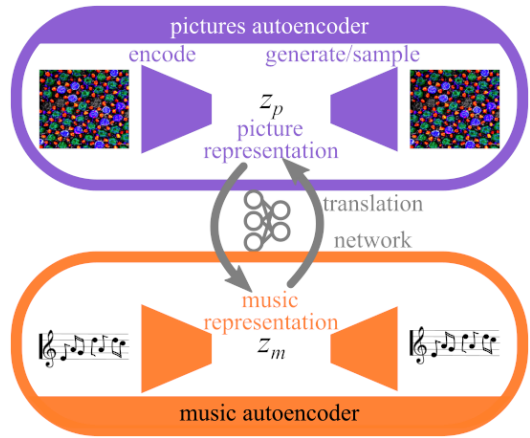}

\caption{Sketch of the encoding/decoding/conversion.}

\end{figure}

\section*{VAE Training}

Autoencoders are neural networks with two constituent parts (Figure
1). An \emph{encoder} recognizes important information in the input,
and compresses it in an internal representation. A \emph{decoder}
reconstructs the most likely original input from a given compressed
representation. Encoder and decoder are trained together, with the
goal of compressing and reconstructing the training inputs. Variational
Autoencoders (VAEs) \citet{kingma_auto-encoding_2013,higgins_beta-vae:_2017}
further assume a continuous representation space, with a Gaussian
prior. Therefore, VAEs can also generate new samples, by remixing
features they encountered in training samples. This generative power
makes VAEs particularly useful for both image and music generation
\citet{roche_autoencoders_2018,broad_autoencoding_2017,roberts_hierarchical_2018}.

The image VAE was trained on a set comprising over 4000 abstract artworks
from several abstract styles in the WikiArt database \citet{wikiart_wikiart.org_nodate},
plus 10000 images generated algorithmically from music samples (see
Online Supplement \citet{wienand_online_nodate}). All images were
downsampled to 64\texttimes 64 pixels for computational manageability.
To represent and generate music, we use Google\textquoteright s pre-trained
MusicVAE \citet{roberts_hierarchical_2018,roberts_musicvae_2018}.
Particularly, the models trained on 2- and 16-bar-long, single-voice
MIDI melodies. Though MIDI is musically limiting, it allows us to
work with melodies and build on clearer, existing translation processes
(see below), which would be harder with natural sound.

\section*{Learning synesthetic associations}

The translation between images and music has at its core a set of
synesthetic associations between visual and musical information. Instead
of leaving this key component entirely to the idiosyncrasies of the
neural networks (as in \citet{muller-eberstein_translating_2019}),
we decided to ground it in a simple algorithmic conversion, based
on \emph{Atomare Klangwelte} \citet{heckl_atomare_2006}. Inspired
by scanning microscopes, the algorithm reads the image row by row,
turning each pixel into a note. Conversely, it also reads music note
by note and renders it in pixel sequences, with longer notes becoming
longer strings of pixels.

\begin{figure}[tbh]
\includegraphics{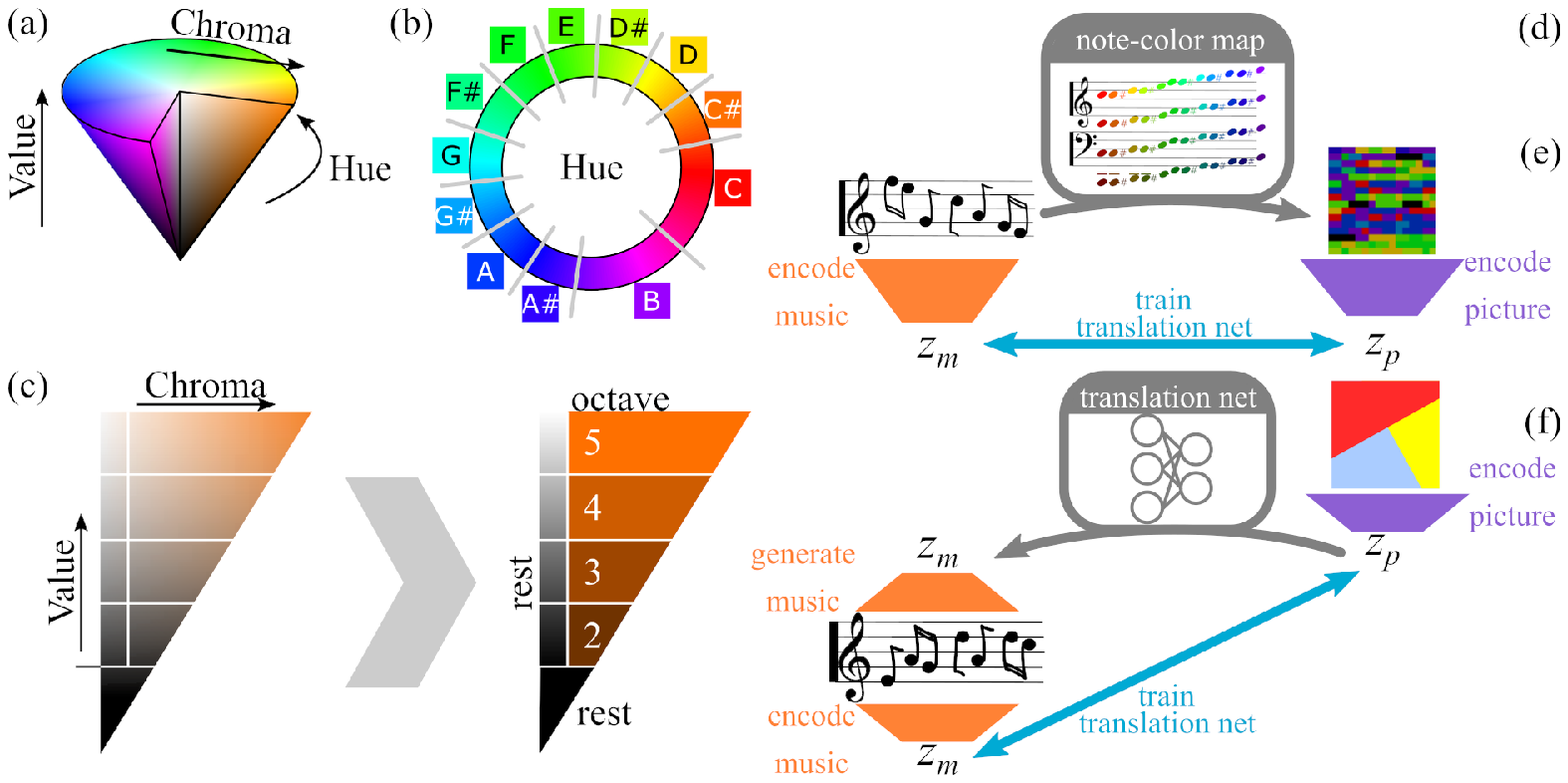}

\caption{Definition of the note-color map and training for the neural network
system\citet{rus_hcl-hcv_2010}.}

\end{figure}

Technical limitations restrict our choice of translation rules. MusicVAE,
for one, works within strictly limited music parameters: it requires
single-voice MIDI samples, all with the same tempo and length, and
notes of constant velocity (i.e., volume). Therefore, notes are reduced
to pitch and duration. To augment the data in notes, we express each
pitch as octave and note on the chromatic scale.

For pixels, we limit information loss by working in color (instead
of grayscale as in \citet{heckl_atomare_2006}). We express each pixel\textquoteright s
color in the Hue-Chroma-Value space (Figure 2(a)) with Chroma (similar
to saturation) fixed. The note-color mapping combines physics-inspired
and shared human synesthetic correspondences \citet{parise_audiovisual_2013}.
Specifically, we associate the color\textquoteright s Hue with a note
on the chromatic scale based on wave frequencies (Figure 2(b)): from
a red C (lowest sound frequency on the chromatic scale and lowest
frequency of visible light), to a violet B (highest sound and visible
light frequency. Though not immediate for humans, we imagine this
numerical link can be an intuitive audiovisual correspondence for
the machine. We also map the color\textquoteright s Value (essentially,
the luminosity) to the note\textquoteright s octave (between C2 and
B5, Figure 2(c)), following the correspondence between low pitches
and darkness, common to humans too \citet{parise_audiovisual_2013}.
This map (incidentally similar to \citet{corra_abstract_1912,castro_jidiji:_nodate})
forms the basis of the perceptual associations of our networks, learned
through repeated exposure and ingrained in the translation network.

\subsection*{Translation network training}

We use the note-color map to turn 10000 MusicVAE-generated melodies
into images (Figure 2(d), which were also added to the training of
the image VAE). The respective encoders compress each melody and image
to their representations. These music-image pairs in representation
space serve as ground truth to train simple multilayer perceptron
networks connecting the two representation spaces (Figures 1 and 2(e)).
Thus the translation networks learn the synesthetic correspondences
implicitly---connecting representation features, not pixels and notes.
Furthermore, the correspondences originate from connections between
neural networks and repeated joint exposure, reflecting some hypotheses
on the origin of human synesthesia \citet{parise_audiovisual_2013,ramachandran_synaesthesia_2001}.

To reinforce the correspondences, we use simplified images to further
train the translation network. From each of WikiArt\textquoteright s
\textquotedblleft Color Field Painting\textquotedblright{} and \textquotedblleft Hard-Edge
Painting\textquotedblright{} styles, 50 works were selected at random
and broken into 64x64 tiles (for a total of almost 10000 tiles). Each
tile is a segment of its original, unmodified picture. Therefore,
putting all tiles from the same picture side by side would reconstruct
the original picture. These tiles are thuse reduced-information versions
of real-world examples (instead of the synthetic images used previously).
The translation network converts them to melodies. Encoding tiles
and melodies produces new music-image pairs, serving as ground truth
to train the translation network again, Figure 2(f). This reinforces
the correspondences and expands the range of representations for which
the translation network learned definite associations. The entire
training process was carried out twice: once based using the 16-bar
MusicVAE, once using the 2-bar version.

From this process emerges an \textquotedblleft artificial synesthete\textquotedblright{}
that is inspired from the color-note associations instead of bound
to them.

\section*{Playing images, painting music}

Figure 3 shows some examples of the artificial synesthete\textquoteright s
work. Since MusicVAE only generates single-voice MIDI, the music samples
focus entirely on the central melodic motif. Image generation is the
opposite. In its pictures, the synesthete conveys blurry impressions,
with little details to focus on.

We made the synesthete generate 16-bar melodies from the images in
the top row of Figure 3(a) (MIDI samples and sheet music in the Online
Supplement \citet{wienand_online_nodate}). The samples clearly show
that the synesthete bends the MusicVAE generator to fit its experience:
melodies are more dissonant, have fewer rests, and less rhythmic variation
than the typical MusicVAE samples (see Online Supplement). The melody
composed from the first sketch on the left obsessively repeats the
same few notes, reflecting the few colors in the image. However, these
are very short notes (mostly 16th), so the networks seem to recognize
the elementary composition of the sketch, but see small chunks of
different shades instead of a large, flat field. Indeed, neural networks
are known to perceive structures and patterns in the pixels that are
invisible to humans \citet{goodfellow_explaining_2015}. The second
sketch has a clearer structure (two equal rectangles), which translates
to a rhythmic structure (roughly every third bar is entirely quarter
notes). Similarly, the third sketch inspires a melody clearly divided
in three parts: beginning with short notes gradually slowing to all
quarter notes. Translating the fourth, darker sketch, the synesthete
clearly perceives lower shades and pitches, and even rests. Finally,
consider Lipide. Much of the detail is lost to the networks (see the
autoencoder\textquoteright s reconstruction in the middle row), which
see a grey-blue haze with few details emerging. Accordingly, the synesthete
generates a melody anchored to an A (blue in the map). Small musical
figures briefly appear before melting back like objects in the fog---or
the details in the reconstructed picture. The inverse process---translating
these melodies back to images---produces the pictures at the bottom
of Figure 3(a). These are visually similar to those obtained from
a simple pass through the image autoencoder (middle row), despite
some information loss. In other words, the translation is consistently
reversible.

\begin{figure}[tbh]
\includegraphics{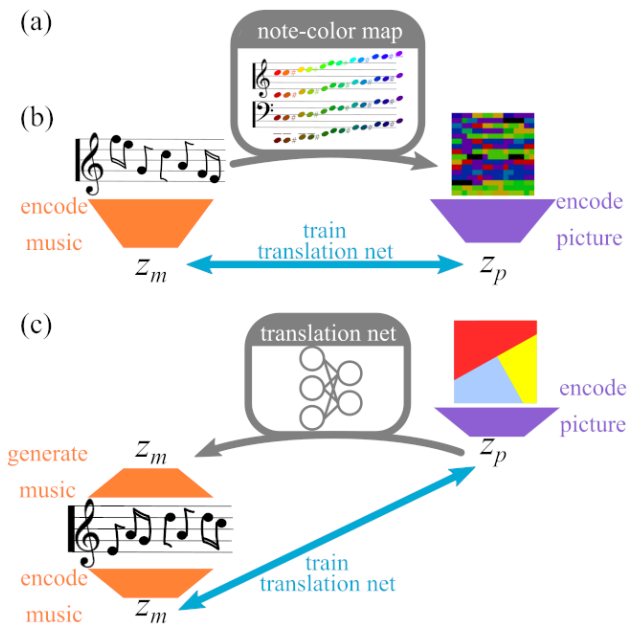}

\caption{Translation examples. (a) Top: original images; middle: autoencoded
reconstructions; bottom: image-music-image reconstructions. (b) Translations
of samples from Bach\textquoteright s \emph{Prelude No.1 in C Maj}
(top) and Beethoven\textquoteright s \emph{Waldstein Sonata} (bottom).
(c) Heterogeneity in latent representation of music (left) and images
(right) for the same samples of Bach (orange) and Beethoven (blue)
as in panel (b).}

\end{figure}

The synesthete was also made to translate samples of well-known classical
music to images, generating series with clearly different flavors
for different pieces, depicted in Figure 3(b) (MIDI samples and sheet
music in the Online Supplement \citet{wienand_online_nodate}). Altogether,
lighter images mirror higher-pitched samples, and one can recognize
some composition elements. Bach\textquoteright s\emph{ Prelude No.1
in C Maj} BWV 846 (top row) has mostly yellow-orange tints with green
features, structurally quite similar to each other. Correspondingly,
the music samples all present the same repeating structure. Beethoven\textquoteright s
\emph{Waldstein Sonata} Op. 53 (bottom row), skews blue. Brown or
reddish features and diversified structures mirror the diverse melodies
(in particular the second and fourth sample).

Figure 3(c) quantifies the diversity of music and image (calculation
details in the Online Supplement). The bars indicate the average normalized
distance in representation space between the samples in Figure 3(b)
and the average of the corresponding series (orange for Bach\textquoteright s
\emph{Prelude}, blue for Beethoven\textquoteright s \emph{Waldstein}).
The \emph{Waldstein} samples have higher heterogeneity than those
from the \emph{Prelude} (which have, for example, a strict rhythmic
structure), and so are the translated picture series. This means that
MusicVAE sees the \emph{Prelude} samples as more similar to each other
than the \emph{Waldstein} ones. Therefore, it encodes them to a smaller
region in the representation space. Analogously, according to the
image VAE, the pictures series obtained from the \emph{Prelude} is
less heterogeneous than that from the \emph{Waldstein}.

\subsection*{Interpolation videos}

Variational autoencoders (VAEs) cab also interpolate between samples,
generating intermediate image and music \citet{wienand_online_nodate,roberts_musicvae_2018}.
We leverage this capability to produce suggestive video sequences.
After picking two images, we had the synesthete translate each to
a 2-bar melody. Using a spherical interpolation between the encoded
representations of the melody extremes, we generated 7 interpolating
music samples (enough to perceive a gradual change, but keeping a
limited cumulative duration). The concatenation of these melodies
gives the audio track. Analogously, we obtain the video track from
the series of images (24 samples per second of music) obtained interpolating
between the starting and ending pictures. Figure 4 shows a sketch
of the process, as well as example bars and frames from the video
in the Online Supplement \citet{wienand_online_nodate}. Visually,
the top and bottom of Nassos Daphnis\textquoteright{} \emph{11-68}
slowly morph into two separate circles. Simultaneously, we hear the
rapid 16th notes become longer and more staccato. Gradually, the pitch
shifts down, until the last two bars, which represent a detail from
Heckl\textquoteright s \emph{Coronen Molek�le}.

\begin{figure}[tbh]
\includegraphics{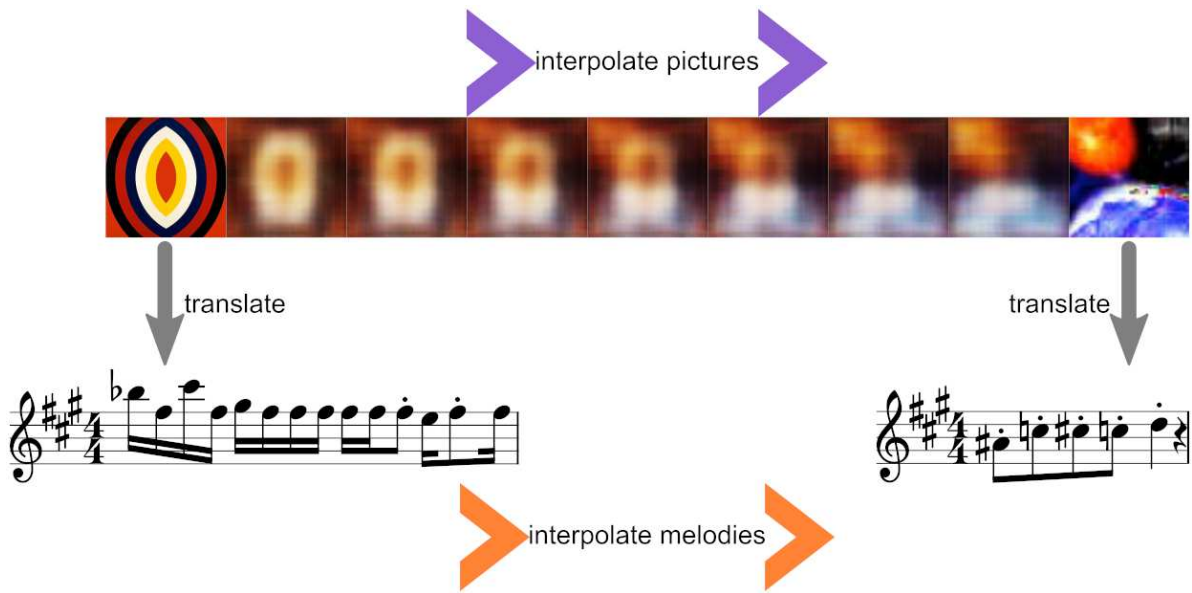}

\caption{Frames from interpolation video and summary of the process. The starting
and ending point are represented with the original images.}

\end{figure}

\section*{Conclusion: Towards a more creative synesthete}

We developed an \textquotedblleft artificial synesthete\textquotedblright{}
that translates between pictures and music. Using variational autoencoders
(VAEs)---a type of neural network---the machine learns to read and
organize visual and musical information. Its synesthetic ability is
rooted in a set of learned correspondences between images and melodies.
These correspondences are the synesthete\textquoteright s interpretation
of an underlying note-color map. Similar learning experience are thought
to be the base of some widely shared cross-modal correspondences in
humans \citet{parise_audiovisual_2013,ramachandran_synaesthesia_2001,maurer_shape_2006}.
The resulting translations are novel works, instead of transposed
data.

The learned correspondences set our artificial synesthete apart from
algorithmic data transposition \citet{castro_jidiji:_nodate,heckl_atomare_2006}
and idiosyncratic translation networks \citet{muller-eberstein_translating_2019},
as they allow us insight to comprehend what the synesthete sees. Moreover,
they increase the creativity of the machine, at least in the framework
of Colton\textquoteright s simple \textquotedblleft creativity tripod\textquotedblright{}
\citet{colton_creativity_2008} (made of appreciation, imagination,
and skill). In fact, these correspondences give the machine better
appreciation. Based on these guidelines, the synesthete decides what
is more or less important to translate, and what are better or worse
translation choices. The weakest leg of the tripod for our synesthete
is certainly imagination. By construction, VAEs remix features they
have seen: they cannot generate fully novel works. Training the image
VAE on a larger and more diverse set of images would give the synesthete
a broader representation space. This would at least increased the
perception of imagination, although the network would remain a remixer.
Finally, our system still has limited generation skill: the images
are still very blurry, the music feels flat. Larger training samples
could improve this aspect as well. Alternatively, generative adversarial
networks have shown integration potential with VAEs for improved image
generation \citet{broad_autoencoding_2017,larsen_autoencoding_2016}.
Music skill could be improved similarly, layering further networks
that articulate from the MusicVAE-generated samples, e.g. varying
tempo and intensity, adding voices or harmony.

Finally, emotions play an important role in color-music correspondences
for humans (synesthete and not \citet{curwen_music-colour_2018,palmer_musiccolor_2013,palmer_music-color_2016})
but not in our neural networks. Affective tagging, already in us to
analyze and generate images and music \citet{alvarez-melis_emotional_2017,you_building_2016,alameda-pineda_recognizing_2016,mohammad_wikiart_2018,ehrlich_closed-loop_2019},
could bridge this gap. However, the unemotional experience of the
artificial synesthete can, by contrast, highlight our own emotions,
as well as elicit new ones when it presents translations we find unexpected
or discordant. This also spotlights the deeply personal nature of
all perception, synesthetic and not. Comparing and contrasting with
the machine\textquoteright s synthetic perception helps us explore
our own, organic experience of music, image, and their pairing.

\bibliographystyle{plainnat}
\bibliography{arxiv}

\end{document}